# *Forging Emotions*: a deep learning experiment on emotions and art

**Amalia Foka**
University of Ioannina

Date of submission: July 2022
Accepted in: January 2023
Published in: January 2023

### Abstract

Affective computing is an interdisciplinary field that studies computational methods that relate to or influence emotion. These methods have been applied to interactive media artworks, but they have focused on affect detection rather than affect generation. For affect generation, computationally creative methods need to be explored that have recently been driven by the use of Generative Adversarial Networks (GANs), a deep learning method. The experiment presented in this paper, *Forging Emotions*, explores the use of visual emotion datasets and the working processes of GANs for visual affect generation, that is, for generating images that can convey or trigger specified emotions. This experiment concludes that the methodology used so far by computer science researchers to build image datasets for describing high-level concepts such as emotions is insufficient and proposes utilizing emotional networks of associations according to psychology research. *Forging Emotions* also concludes that to generate affect visually, merely corresponding to basic psychology findings, such as bright or dark colours, does not seem adequate. Therefore, research efforts should aim to understand the structure of trained GANs and compositional GANs in order to produce genuinely novel compositions that can convey or trigger emotions through the subject matter of generated images.

## Forging Emotions*: un experimento de aprendizaje profundo sobre las emociones y el arte*

**Resumen**

*La computación afectiva es un campo interdisciplinario que estudia los métodos computacionales que se relacionan o influyen en la emoción. Estos métodos se han aplicado a las obras de arte de los medios interactivos, pero se han centrado en la detección de afectos en lugar de su generación. Para la generación de afectos, se tienen que explorar los métodos computacionalmente creativos, los cuales últimamente han sido impulsados por el uso de redes generativas antagónicas (GAN en inglés), un método de aprendizaje profundo. El experimento presentado en este trabajo,* Forging Emotions *(Forjando Emociones), explora el uso de conjuntos de datos de emociones visuales y los procesos de trabajo de las GAN para la generación de afectos visuales, es decir, para generar imágenes que puedan transmitir o desencadenar emociones específicas. Este experimento concluye que la metodología utilizada hasta ahora por los investigadores de ciencias de la computación para construir conjuntos de datos de imágenes que describan conceptos de alto nivel como las emociones es insuficiente y propone utilizar redes emocionales de asociaciones de acuerdo con la investigación en psicología.* Forging Emotions *también concluye que, para generar afectos de manera visual, no parece adecuado limitarse a los hallazgos básicos de la psicología, por ejemplo, los colores brillantes u oscuros. Por lo tanto, los esfuerzos de investigación tienen que dirigirse a comprender la estructura de las GAN formadas y las GAN compositivas para producir composiciones genuinamente nuevas que puedan transmitir o desencadenar emociones por medio del tema de las imágenes generadas.*

## Introduction

The notion of emotion, as we understand it today, was established through scientific research in various disciplines. Emotions have been studied from the perspective of various disciplines, including psychology, neuroscience, sociology, medicine, history as well as computer science. As a result, the question "What is an emotion?" will rarely generate the same answer from different scientists or individuals. This is primarily due to the different basis of the emotional phenomena or theoretical issues studied from every discipline. The main questions asked regarding emotions include: where they originate, what physiological, behavioural, and cognitive changes they produce, what different expressions (e.g., facial) they induce, and what subjectively experienced feelings are.

Katja Kwastek (2013) has studied the aesthetic experience of the recipient in the case of interactive media artworks, which often involves the arousal of emotions. In non-interactive works, the recipient's emotional experience is staged in the past tense; an artist either represents the emotions he/she has experienced or plans the desired emotional experience for the recipient, albeit mostly subjective. In contrast, interactive media artworks are situated in the present tense. However, as Kwastek notes, even if the interaction process leaves scope for the unexpected, the orchestration of emotional experiences is designed in an even more explicit manner, since feedback processes must be programmed into the technical system implementing the artwork.

Consequently, affecting computing methods become extremely pertinent to the orchestration of emotional experiences in interactive media artworks. However, affective computing research is largely concentrated on detecting emotions rather than on generating or triggering them. Accordingly, interactive media artworks that consider affective computing methods also mainly utilize them for affect detection.

Given the problem of affect generation, computational creativity methods need to be explored, which most commonly involve Generative Adversarial Networks (GANs), a deep learning method. Therefore, when considering the generative process of GANs, and more importantly, when they are intended to generate affect, the utilized training image datasets are of crucial importance.

The experiment presented in this paper, *Forging Emotions*, explores the use of visual emotion datasets and the working processes of GANs for visual affect generation: that is, for generating images that can convey or trigger specified emotions. It appears that researchers attempting to build visual emotion datasets have almost reached a consensus on the most appropriate methodology for that purpose. This methodology consists of searching social media for tags with words relating to emotions (e.g., anger, disgust, fear, sadness) and collecting images. A validation procedure is then performed wherein a large number of people confirm whether or not the emotion tags are correct. *Forging Emotions* aims to confront the idea that searching social media with a single word can result in a set of images that truly represent what an emotion is. Moreover, it aims to question the validation procedure utilized to account for the subjectivity of emotions. Finally, it aims to explore the working processes of GANs for visual affect generation and which visual aspects of the generated images are able to convey or trigger specified emotions.

## 1. Background

### 1.1. Affective computing

Research in affective computing is mainly focused on building systems that can respond and adapt to human emotions (Calvo *et al.* 2015). Such systems first need to be able to detect human emotions. For this reason, a great variety of sensors have been developed that can monitor physiological signs that are later utilized for recognizing emotions. The modalities of facial expressions, body gestures, and speech have also been used for this purpose.

When detecting emotions, Paul Ekman's basic emotion theory (Ekman 2005) is the most commonly used. It suggests that all emotions derive from a limited set of universal and innate basic emotions. The following six basic emotions are considered: anger, disgust, fear, happiness, sadness, and surprise. It is also common for a seventh, neutral emotional state to be added. Additionally, there are cases where a few other emotions are considered, such as anxiety, excitement, calm, and lust, among others. More recently, two-dimensional models of emotion have attracted interest. One very commonly used two-dimensional emotion model is the valence and arousal model (Eerola & Vuoskoski 2011). This proposes that all affective states can be understood as varying degrees of valence (a pleasure–displeasure continuum) and arousal (activation–deactivation).

To a large extent, affective computing research focuses on detecting rather than on generating or triggering emotions. A considerable body of research on affect generation involves synthesizing human-like expressions of emotion through facial features, speech, and gestures for virtual characters and robots (Calvo *et al.* 2015). Affective games are a sub-field of affective computing studying how gameplay can be adapted according to players' emotions and how emotions can be triggered (Lara-Cabrera & Camacho 2019). Nevertheless, current affective games adapt to or trigger emotions by design rather than with computational methods.

In conclusion, research on generating affective triggers is largely unexplored. Accordingly, interactive media artworks that use affective computing methods also mainly consider affect detection rather than affect generation.

For example, the work *Chameleon* (2008/2010) by Tina Gonsalves recognized the audience's emotional state using a facial expression detection system and classified it as one of the six basic emotions as defined by Ekman. Then, the detected emotions were sent to the video engine, which triggered videos from the new corpus Gonsalves created to empathize with the audience's emotional state (Gonsalves, Berthouze & Iacobini 2010). Gonsalves created her own facial emotion dataset that was "more dynamic and aesthetic, engaging and emotionally probing". Gonsalves' experience in working with affective computing methods and visual emotion datasets led her to the conclusion that emotions expressed and monitored in laboratories for scientific research "don't often correlate to the emotions that form the fabric of our everyday lives". She also notes: "There seem to be limited emotions being explored, visually underwhelming databases being used, and the non-ecological settings such as the lab to test responses […] using small groups of subjects with narrow representation, what does the knowledge that science is building about emotions actually mean?".

### 1.2. Deep learning methods

Computational Creativity is a field that has lately attracted much new interest with the advent of Generative Adversarial Networks (GANs) (Goodfellow *et al.* 2014), a deep learning method. If the problem of affect generation is to be considered, then the processes involved in GANs have to be explored. In music, there is the field of affective algorithmic composition (AAC) (Williams *et al.* 2015), where an intended affective response always informs the algorithm for music composition. There are many successful applications of GANs within the field of AAC. However, visual affect generation, that is, the generation of new images that intend to convey or trigger specific emotions, is still unexplored.

GANs have already been explored by artists such as Anna Ridler (2017) and Mario Klingemann (2019), but not for visual affect generation. One of the very few works in the literature that considers the generation of artworks that convey a specifically defined emotion is that of David Alvarez-Melis and Judith Amores (2017). For this work, a GAN was trained on 13,000 images of emotion-labelled modern art paintings. Alvarez-Melis and Amores concluded that their approach was able to generate artworks with high-level emotional features that agree with psychology literature; for example, red for anger, dark colours for sadness and fear, and images that resemble natural landscapes for joy. GANs have already exhibited the ability to generate images that are hard to distinguish from "real" ones. For example, one of the most well-known GANs was developed by NVIDIA for generating new face images that are hard to identify as images not of real people (Karras, Laine & Aila 2019). Additionally, approaches to synthesizing facial affect have also been presented in the literature (Park, Kim, & Ro 2019).

However, in most cases, GANs have successfully generated images of objects or concepts that are quite specific in form and structure, such as human faces, dogs, landscapes, and others. As a result, the training datasets used comprise very similar images. For example, the NVIDIA training dataset includes only photographs of human faces that are also automatically aligned and cropped.

Nevertheless, when higher-level concepts are considered, such as the whole notion of emotions without being limited to facial expressions, the question remains of how to create a suitable training image dataset that could be later used with GANs to generate new affective images, that is, those that convey or trigger a specified emotion.

### 1.3. Visual emotion datasets

Most datasets used for visual emotion analysis are targeted toward applications for affect detection. However, even if the problem of visual affect generation is considered, a similar visual emotion dataset is also required in order for the system to learn how to generate affective images. For this purpose, the most commonly used visual emotion datasets are summarized below. It should be noted that this discussion will not include datasets of facial expressions that are beyond the scope of the presented experiment, *Forging Emotions*.

Emotion researchers have constructed a number of visual emotion datasets. They commonly define specific categories of images according to their research questions and then hand-select images. For example, the International Affective Picture System (IAPS) was designed to provide a standardized set of pictures for studying emotion and attention and was developed by the National Institute of Mental Health Center for Emotion and Attention at the University of Florida (Lang, Bradley & Cuthbert 2005). IAPS is the most widely used visual emotion dataset in hundreds of behavioural and neuroimaging studies. It includes over 1,000 images depicting people experiencing various emotions (e.g., sad, fearful, angry), erotic couples, funerals, dirty toilets, cityscapes, landscapes, wars and disasters, mutilated bodies, baby animals, and many more. Other visual emotion datasets constructed by emotion researchers include the Geneva Affective Picture Database (GAPED) (Dan-Glauser and Scherer 2011) with 730 images and the Nencki Affective Picture System (NAPS) (Marchewka *et al.* 2014) with 1,256. In the GAPED dataset, images of spiders and snakes and scenes of moral or legal violations were selected as negative images, while mainly humans, animal babies, and landscapes were used as positive images. Images in the NAPS dataset belong to five broad categories: people, faces, animals, objects, and landscapes. The "object" category is a broad one including a wide range of clearly visible objects, foods, or vehicles depicted without humans or animals. Pictures in all categories included stimuli for different emotions. For example, images in the "people" category included live, injured, or dead human bodies. In all the aforementioned visual emotion datasets, images were evaluated on valence and arousal.

Computer scientists also have built visual emotion datasets. Mikels *et al.* (2005) created the Subset A of IAPS (IAPSa) by selecting 395 pictures from IAPS and then categorizing them into eight discrete categories (anger, disgust, fear, sadness, amusement, awe, contentment, and excitement) by conducting a user study. Machajdik and Hanbury (2010) created another dataset of 807 artistic photographs downloaded from DeviantArt, an online social community for artists and art enthusiasts. The same eight basic emotion categories were also used for this dataset. The associated emotion for each image was the label given by its owner on DeviantArt. A more recent dataset is the Open Affective Standardized Image Set (OASIS) (Kurdi, Lozano, & Banaji 2017). It contains 900 images, collected using the Google Images search engine, depicting a broad spectrum of themes, including humans, animals, objects, and scenes. Each image was rated on valence and arousal by recruiting participants through Amazon's Mechanical Turk (MTurk).

You *et al.* (2016) aimed to create a large visual emotion dataset and to highlight that all the previously mentioned datasets are significantly smaller than those used for other computer vision tasks. For example, the ImageNet dataset (Deng et al. 2009) used for object recognition as well as image generation with GANs contains more than 14 million images. Hence, You et al. collected over 3 million images from Flickr and Instagram labelled with one of the eight basic emotions. They also used MTurk to verify the emotion labels associated with images. The created dataset ultimately contains a total of over 23,000 images with verified emotion labels: currently the largest visual emotion dataset available.

This discussion surrounding visual emotion datasets can conclude that computer science researchers generally turn to online resources, particularly social media, to collect images associated with emotions with single-word queries. Then, most commonly, MTurk is used so that many different people can validate the emotion conveyed or triggered by each image. Thus, it seems that researchers in the field have almost reached a consensus on the most appropriate methodology, and what remains to be completed is the construction of a vast visual emotion dataset containing millions of images.

## 2. *Forging Emotions*

It is uncertain whether the previously described visual emotion datasets can genuinely visualize how emotions are experienced or evoked. An emotion is not just a word, such as the ones used to search social media, but a network of associations. According to psychology research, each individual relates an emotional percept or event in many different ways to a multitude of past emotional experiences (Fellous & Robinson 2006). The basic idea is that, for example, the emotion of happiness is not just a word but rather a concept that a network with nodes of past emotional experiences can represent. Thus, the network of happiness could include nodes for the emotional experience of a child being born, a vacation, being with friends and family, and many more. Everyone experiences emotions differently, and thus emotional networks of associations are individual; each individual may have different emotional experiences related to happiness. However, according to psychology literature, emotional networks share a common basic structure for most people (Fellous & Robinson 2006).

*Forging Emotions* aims to confront the idea that searching social media with a single word can result in images that genuinely visualize how emotions are experienced or evoked. Moreover, it questions the process of employing MTurk to account for the subjectivity of emotions. For this purpose, two new image datasets were created by searching Instagram for the hashtags #sad and #happy. The collected images were not verified in any way.

None of the previously mentioned visual emotion datasets consider the emotion of happiness, even though it was included in the six basic emotions originally introduced by Ekman. In social media behaviour studies, it is often concluded that one always tries to appear attractive, happy, and clever (Freitas 2017). In the *Forging Emotions* experiment context, it seems paradoxical not to explore the "happy" emotion when social media is being used to understand emotions. The #happy dataset poses the question of whether this is an emotion that can be better understood from social media data or if shared emotional experiences on social media are authentic.

This experiment also aims to test whether new images can be composed that visualize and evoke the emotions of happiness and sadness, or can in some way "forge emotions", which relates to commonly naming the generator component of a GAN a "forger." For this reason, the experiment was named *Forging Emotions*.

GANs have been successfully applied to the generation of new images of human faces, dogs, landscapes, and more. However, criticism of the use of GANs mentions that they are not interpretable and are often considered black boxes. Consequently, interpreting and understanding what a GAN has learned is an active research topic within Explainable Artificial Intelligence (XAI) (Xu *et al.* 2019), which has attracted much recent interest. For example, the recent work of Bau *et al.* (2018) aimed to understand whether a GAN learns composition or merely memorizes pixel patterns. The conclusion was that GANs indeed learn aspects of composition and that certain neuron units have learned specific features of the taught domain.

Another exploration strategy for interpreting and understanding deep learning models is black-box exploration. In this type of exploration, only the training dataset and the deep learning model's output are used to examine its behaviour and provide insight into its interpretations (Wang 2019). This approach is similar to the one Ridler (2017) describes, in which she used GAN images as a mirror to her own drawing process.

In this respect, *Forging Emotions* will also perform a black-box exploration in order to understand what a GAN has learned when trained with a visual emotion dataset.

## 3. Experiments

The experiments were performed with the implementation, and the proposed architecture guidelines, for stable Deep Convolutional GANs (DCGANs) (Radford, Metz & Chintala 2015). The training images were not pre-processed, except by scaling them to the range of the tanh activation function [-1, 1]. All models were trained with mini-batch stochastic gradient descent (SGD) with a mini-batch size of 64 for 100,000 epochs. All weights were initialized from a zero-centred normal distribution with a standard deviation of 0.02. In the LeakyReLU, the slope of the leak was set to 0.2 in all models. The Adam optimizer was used with a learning rate of 0.0002. Finally, the momentum term $\beta_1$ was set to 0.5. The images were generated at a 64 x 64 pixel resolution. In all the following experiments, the DCGANs were trained with the same parameters and for the same number of epochs.

### 3.1. Dataset creation

Two new datasets were created for performing the following experiments *Forging Sadness* and *Forging Happiness* with the intent of verifying whether the adopted methodology by computer scientists is suitable for visual affect generation.

As previously noted in the discussion of visual emotion datasets, computer scientists generally turn to online resources, particularly social media, to collect images associated with emotions with single-word queries: the name of an emotion. Then, most commonly, MTurk is used so that many different people can validate the emotion conveyed or triggered by each image.

A similar procedure was followed for the creation of the new datasets in order to assess the methodology employed by computer scientists. For this, the simple tool Instaloader v4.9.6 (2023) was used with the running option #hashtag to download images with a certain hashtag. In the case of the experiment *Forging Sadness*, the hashtag #sad was used, whereas in *Forging Happiness*, the hashtag #happy was used. 30,000 unique images were collected for each.

The size of the created datasets is considered appropriate for training a DCGAN to generate new images and avoid the problem of underfitting, since it is significantly larger than any other visual emotion dataset currently available. For example, the currently larger available visual emotion dataset by You *et al.* (2016) contains 2,635 images relating to the emotion of sadness. On the other hand, the emotion of happiness is not included in any of the currently available visual emotion datasets.

Finally, the images collected from Instagram were not processed in any way other than to remove duplicates. Furthermore, the collected images were not verified by humans or in any other way. This aimed to fulfil the purpose of the performed experiments, which is to determine whether the methodology employed by computer scientists for creating visual emotion datasets with single-world queries and human validation is appropriate for visual affect generation. Another purpose of these experiments is to conclude the type and variety of the images users post on Instagram with such hashtags. Eventually, this type of information will establish the issues present in the dataset and finally determine alternative, more suitable methodologies for creating visual emotion datasets.

### 3.2. Forging sadness

Initially, the image dataset created for the *Forging Emotions* experiment, involving 30,000 Instagram images with the hashtag #sad, was used to train the DCGAN. Then, new images were generated; these are shown in figure 1. Next, the same procedure was followed for the dataset created by You *et al.* (2016) for the emotion of sadness, which included 2,635 images. The images generated in this case are shown in figure 2.

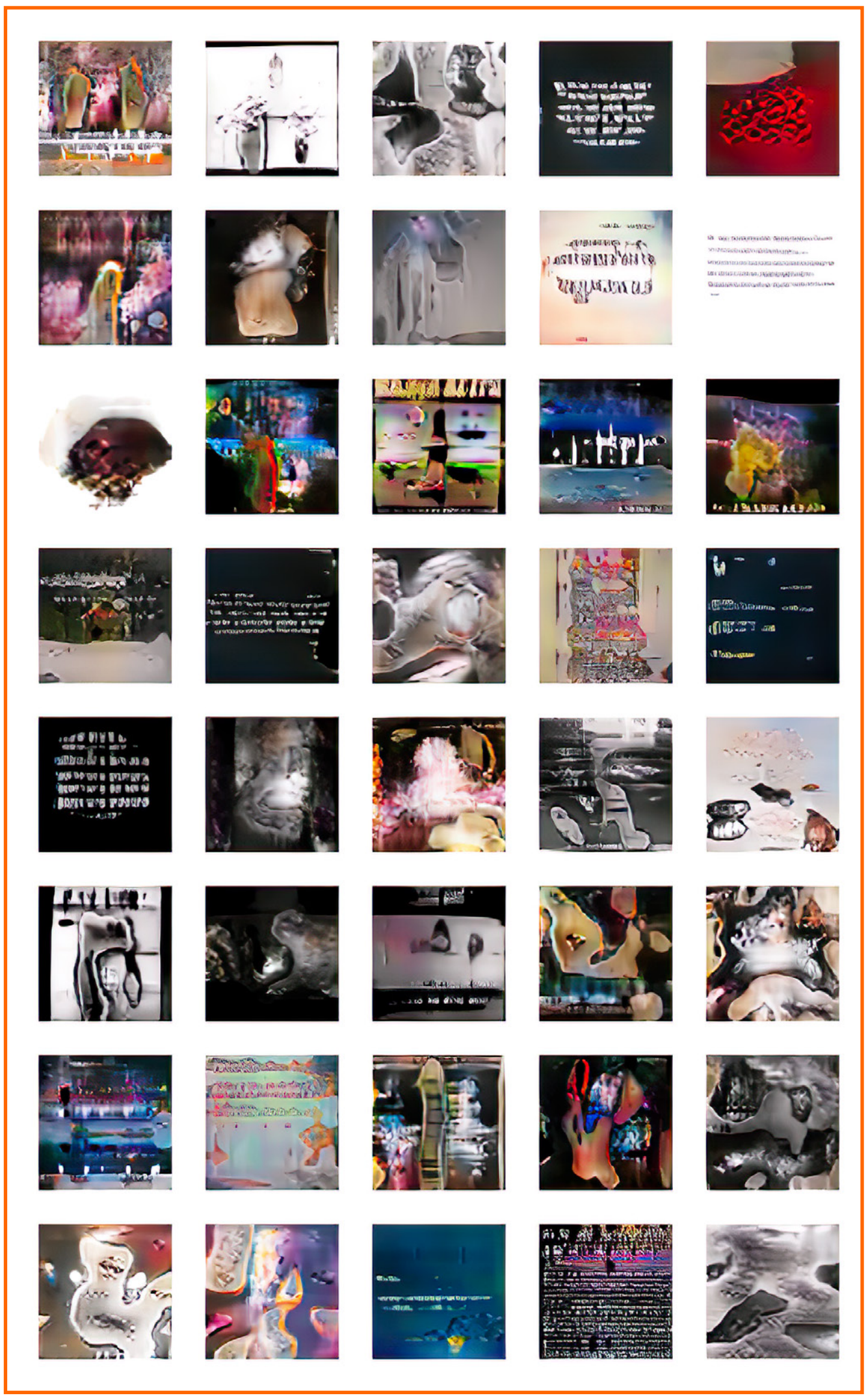

Figure 1. The DCGAN generated images when trained with 30,000 Instagram images with the hashtag #sad. Source: http://www.amaliafoka.com/img/gallery/sad2-upscale2.jpg

Both Figures show more images with dark colours, in line with psychology literature, which generally associates dark colours with unpleasant emotions. The colour statistical analysis confirms that the largest colour cluster in figure 1 is of dark grey (RGB 27, 26, 26), with a total coverage of 26.84% of the pixels in all generated images. Similarly, in figure 2, the largest colour cluster is charcoal (RBG 24, 18, 19), with a total coverage of 30.63% of the pixels in all generated images. The rest of the high percentage colour clusters are for lighter shades of grey. Other colours are present in the generated images of figure 1, with a total coverage of 12.75%, and in figure 2, with a total of 10%. The average RGB values are similar in figures 1 and 2 (116, 108, 108) and (98, 89, 88) respectively.

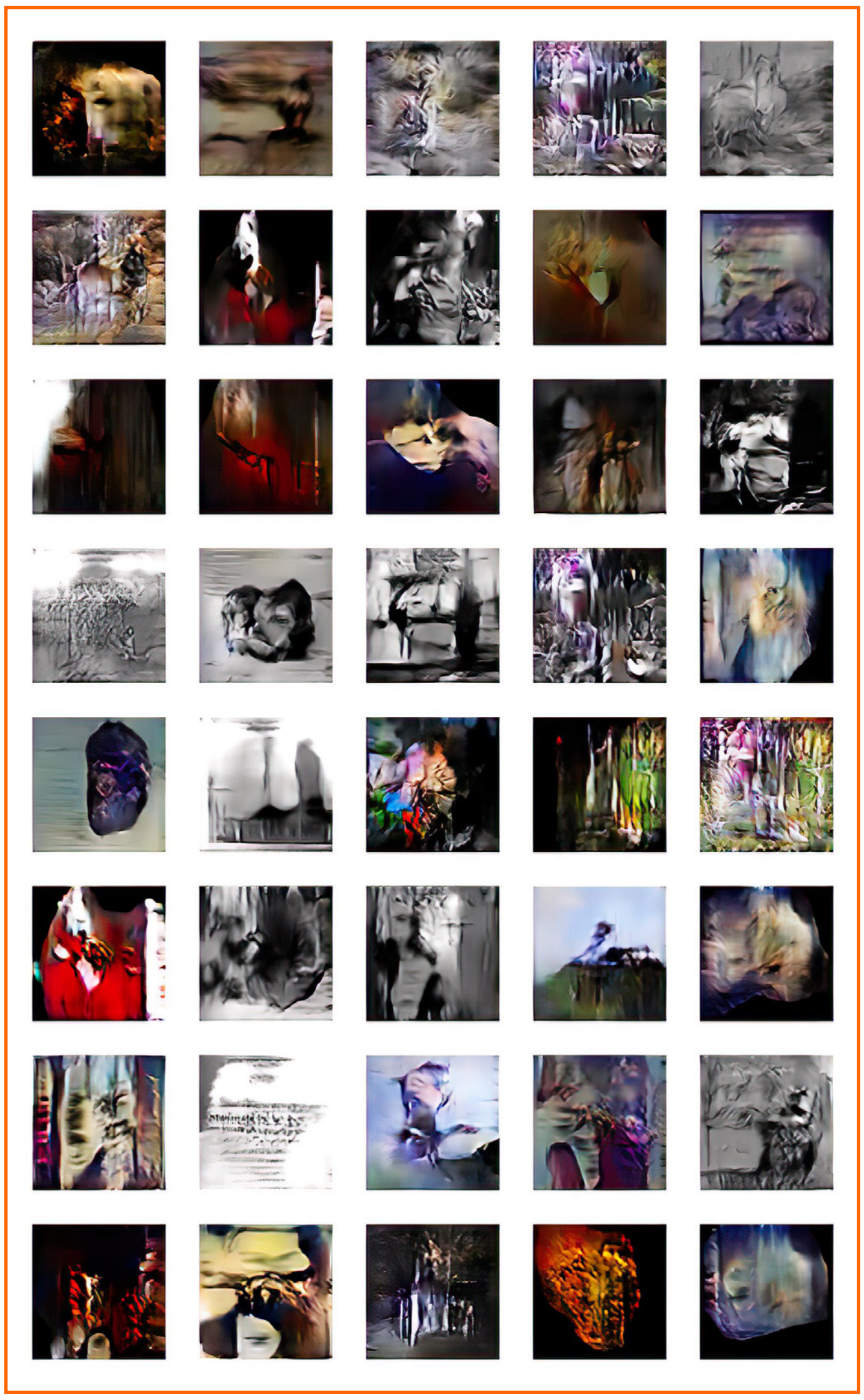

Figure 2. The DCGAN generated images when trained with the image dataset created by You *et al.* for sadness. Source: http://www.amaliafoka.com/img/gallery/sadness-upscale2.jpg

To analyse the generated images' subject matter, the pre-trained ClipCap model (Mokady, Hertz & Bermano 2021) was used to obtain captions. Two captions were obtained for each generated image with the ClipCap model trained on the COCO-captions dataset (Chen *et al.* 2015) and on the Conceptual Captions dataset (Sharma *et al.* 2018). For example, the COCO caption for the image in figure 1 (row 1, column 3) is "A group of urinals mounted to the side of a wall" and the Conceptual caption for the same image is "I'm not sure what kind of shoes this is, but I'm pretty sure it's a pair of shoes". The COCO and Conceptual captions generated for the image in figure 2 (row 8, column 4) are "a close up of a traffic light with a sky background" and "the images were captured by person, who works as a geologist" respectively. One example of two similar captions is the image in figure 2 (row 2, column 4):

"A blurry image of two people kissing in front of a window" and "couple kissing in the dark".

It can be observed that all the generated images shown in both figures are, in some way, reminiscent of abstract paintings. In both cases, some images appear as abstract human figures or, in other cases, abstract landscapes or cityscapes. The obtained captions also validate this. The Conceptual captions "I'm not sure what this is, but it looks like a painting" and "what is the name of this painting?" were obtained for several generated images, for example, in figure 2 for images at (row 3, column 4) and (row 4, column 2). On the other hand, the obtained COCO captions contain most often the characterization of a "blurry photo" or "blurry image" and, in other cases, the characterization of a "collage of pictures".

One distinct difference is that figure 1 includes more images that look like text, due to the different types of images included in the two training datasets used. The captions obtained validate this. For example, the COCO caption for the image in figure 1 (row 1, column 4) is "A black and white photo of a computer screen" whereas the Conceptual caption is "this is what happens when you're in the middle of a text message". The captions for the image in figure 1 (row 2, column 5) are "A blurry image of a computer monitor and keyboard" and "this is a screenshot of the page". In general, the generated images that look like text were most often captioned by COCO as a computer monitor and by Conceptual as a screenshot.

Humans validated the dataset used for generating the images shown in figure 2, which mainly consisted of images depicting humans with facial or body expressions of sadness. The dataset for sadness created by You *et al.* also included a number of images of pets. Other types of images constitute only a tiny proportion of the dataset and include text images, graves or graveyards, dead flowers or plants, winter landscapes, and similar. The You *et al.* dataset (2016) included only 2,635 images relating to the emotion of sadness, and it is evident that a significantly larger dataset is required for the GAN to be able to generate less abstract images.

On the other hand, the image dataset collected for the *Forging Emotions* experiment included predominantly textual images. Our dataset also included many images depicting humans, pets, dead plants, and winter landscapes in addition to images of comics, movie stills, bad food, and other types of images not present in the dataset by You *et al.* (2016). Finally, since the sadness dataset created for the *Forging Emotions* experiment was not validated, it included some images that were not associated with the emotion of sadness but rather with the meaning of the word "sad" for characterizing something as inadequate or unfashionable.

Nevertheless, it becomes evident that collecting social media images with single-word queries cannot guarantee that the whole range of how people experience emotions can be included. For example, we cannot find any images of sadness in either of the two considered datasets that depict, for instance, funerals, wars, disasters, or mutilated bodies, as was the case in the IAPS dataset created by emotion researchers. Moreover, it is most probable that such images would be removed even when posted on social media, due to policies.

Furthermore, the generated captions also fail to infer the emotion of sadness. Captions include the names of the detected object classes, such as people, cars, cats, television, luggage rack, and others, but no descriptions of unpleasant emotions were obtained.

In conclusion, although the employed validation method with MTurk ensures that some irrelevant pictures would be removed from the dataset, it cannot accommodate for the fact that several aspects of how sadness is experienced are missing. Social media users' behaviour and emotion research should be studied further in order to identify multiple image sources and the emotional networks of associations that constitute how emotions are experienced. Finally, significantly larger datasets should be built for training GANs.

### 3.3. Forging Happiness

Another DCGAN was trained using the image dataset created by collecting 30,000 Instagram images with the hashtag #happy, and the generated images are shown in figure 3.

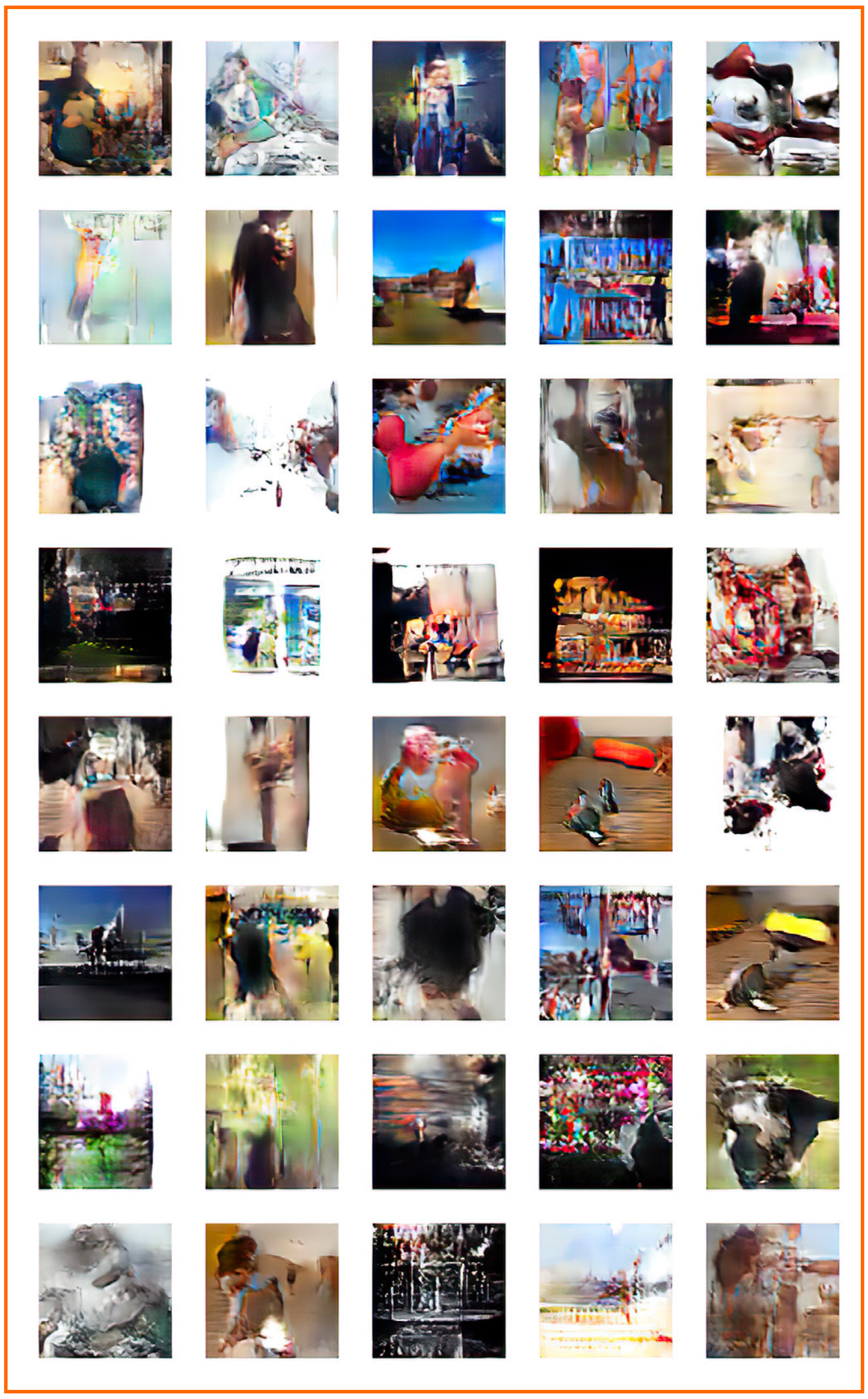

Figure 3. The DCGAN generated images when trained with 30,000 Instagram images with the hashtag #happy. Source: http://www.amaliafoka.com/img/gallery/happy-upscale3.jpg

Again, all generated images are reminiscent of abstract paintings, including abstract human figures, landscapes, and cityscapes. The obtained captions again include characterizations such as "painting", "collage" and "blurry photo".

Images in figure 3 have noticeably brighter colours than those in Figures 1 and 2, which is also in line with psychology literature that generally associates bright colours with pleasant emotions. The colour statistical analysis confirms that the largest colour cluster in figure 3 is porcelain (RGB 220, 221, 218), with a total coverage of 26.28% of the pixels in all generated images. Colours other than white or grey are present in the generated images of figure 3, with a total coverage of 28.53%: almost triple the percentage observed in Figures 1 and 2. The average RGB value for figure 3 is (135, 126, 120) which is a lighter colour value than for Figures 1 and 2.

Instagram users post a greater variety of images when using the hashtag #happy. In addition to the many pictures depicting humans, there are also many featuring luscious food, photos from vacations including landscapes and cityscapes, images showing an object of desire, babies, and more.

Even though the whole range of how people experience happiness is still not included in the collected images, what becomes more evident in this case is that it is questionable that the generated images can visualize or trigger the emotion of happiness. Generating images that merely correspond to basic psychology findings, such as bright colours as in the work of Alvarez-Melis and Amores (2017), does not seem adequate for visual affect generation. For example, although red is generally associated with anger in psychology findings, this is not always the case. In an image depicting a natural landscape at sunset, the sky is often red, and this would trigger positive emotions for many people.

The first two images in the second row of figure 3 quite clearly depict women posing. These images were generated because the #happy dataset included many images posted by women posing with their new clothes or other types of "selfie" images to state that they are happy for various reasons. The Conceptual captions generated for these images are "painting of a woman's face" and "the video shows a woman standing in front of a wall with her arms outstretched and her hands covering her face".

Accordingly, the third image on the second row of figure 3 depicts a natural landscape similar to those often posted on Instagram from vacations. Nevertheless, the COCO and Conceptual captions for this image are "A blurry image of a woman playing a video game" and "the video shows a man standing in front of a blue sky with clouds".

In general, the captions obtained for the generated images in figure 3 sometimes describe occasions related to the emotion of happiness. For example, the Conceptual captions obtained for the images in figure 3 at (row 1, column 4), (row 4, column 3), (row 6, column 2), and (row 6, column 3) describe "people dancing". Another example is the caption "slow motion: the video shows a pair of dolphins playing in the water" for the image in figure 3 (row 1, column 5). Finally, the caption "the wedding of noble person and businessperson was attended by hundreds" for the image at (row 2, column 5) and the caption "funny animals of the week, funny animal photo, cute animal pictures" for the images at (row 5, column 4) and (row 6, column 5).

Even though some relatively successful examples can be found in the experiment for happiness, as exhibited in the *Forging Sadness* experiment, significantly larger datasets should be built for training GANs that include more aspects of how happiness is experienced.

More importantly, recent research efforts into understanding the structure of trained GANs using exploration strategies, such as GAN dissection (Bau *et al.* 2018) and black-box exploration (Wang 2019), could shed light on the different features learned from visual emotion datasets. This knowledge could be applied to training compositional GANs that could create genuinely novel compositions and combine different elements from different types of images. Thus, the subject matter of generated images would be better directed at visually generating affect.

## Conclusion

The experiment *Forging Emotions* was doomed to fail in its initial goal of generating new images that convey or trigger the emotions of happiness and sadness. Nevertheless, its execution allowed us to draw two main conclusions and instigate research into the issues identified. First, the methodology used so far by researchers to build datasets for describing high-level concepts such as emotions is not sufficient. This conclusion is in line with comments made by artists while working on affective computing methods, such as the issue raised by Gonsalves regarding whether the datasets used in scientific research, especially in the case of emotions, can really "correlate to the emotions that form the fabric of our everyday lives". It seems that the collected images based on single-word queries can indeed generally be characterized as related to an emotion, such as happy or sad, but cannot describe the full range of the emotional experience of happiness or sadness. It would appear that a more appropriate methodology would be first to explore a network of associations for an emotion that describes an experience rather than just the name of an emotion. Additionally, a greater variety of online sources should be considered rather than only social media. Finally, the contribution of artists in the creation of visual emotion datasets would be most valuable.

The last conclusion of this experiment is that for visual affect generation, research efforts should aim to better understand the structure of trained GANs and compositional GANs. In many cases, GANs are able to understand the different elements used to compose, for example, a scene. If visual emotion datasets are built according to emotional networks of associations, they will comprehensively visualize how each emotion is experienced. Then, methods and frameworks for understanding and interpreting GANs will shed light on the different features learned from visual emotion datasets. This knowledge could be applied to training GANs that can create genuinely novel compositions and combine different elements from different types of images. In this way, not only the colours but also the subject matter of generated images will be better directed at visually generating affect.

## References

Alvarez-Melis, David and Judith Amores. "The Emotional GAN: Priming Adversarial Generation of Art with Emotion". *31st Conference on Neural Information Processing Systems* (NIPS), 2017.

Bau, David, Jun-Yan Zhu, Hendrik Strobelt, Bolei Zhou, Joshua B. Tenenbaum, William T. Freeman and Antonio Torralba. "GAN Dissection: Visualizing and Understanding Generative Adversarial Networks". *ArXiv* (2018, November). DOI: https://doi.org/10.48550/arXiv.1811.10597

Calvo, Rafael A., Sidney D'Mello, Jonathan M. Gratch and Arvid Kappas (eds.). *The Oxford Handbook of Affective Computing*. Oxford University Press, 2015. DOI: https://doi.org/10.1093/oxfordhb/9780199942237.013.040

Chen, Xinlei, Hao Fang, Tsung-Yi Lin, Ramakrishna Vedantam, Saurabh Gupta, Piotr Dollar and C. Lawrence Zitnick. "Microsoft COCO Captions: Data Collection and Evaluation Server". *ArXiv* (2015, April). DOI: https://doi.org/10.48550/arXiv.1504.00325

Dan-Glauser, Elise S. and Klaus R. Scherer. "The Geneva Affective Picture Database (GAPED): A New 730-Picture Database Focusing on Valence and Normative Significance". *Behavior Research Methods*, vol. 43, no. 2 (2011): 468-477. DOI: https://doi.org/10.3758/s13428-011-0064-1

Deng, Jia, Wei Dong, Richard Socher, Li-Jia Li, Kai Li and Li Fei-Fei. "ImageNet: A large-scale hierarchical image database". *2009 IEEE Conference on Computer Vision and Pattern Recognition. IEE* (2009): 248-255. DOI: https://doi.org/10.1109/CVPR.2009.5206848

Eerola, Tuomas and Jonna K. Vuoskoski. "A comparison of the discrete and dimensional models of emotion in music." *Psychology of Music*, vol. 39, no. 1 (2011): 18-49. DOI: https://doi.org/10.1177/0305735610362821

Ekman, Paul. "Basic Emotions". *Handbook of Cognition and Emotion* (Chichester, UK: John Wiley & Sons, Ltd, 2005), 45-60. DOI: https://doi.org/10.1002/0470013494.ch3

Fellous, Jean-Marc and Jenefer Robinson. "A Mechanistic View of the Expression and Experience of Emotion in the Arts". *The American Journal of Psychology*, vol. 119, no. 4 (2006): 668. DOI: https://doi.org/10.2307/20445371

Freitas, Donna. The Happiness Effect: How Social Media Is Driving a Generation to Appear Perfect at Any Cost. Oxford University Press, 2017.

Gonsalves, Tina, Nadia Berthouze and Matt Iacobini. "The Chameleon Project". *Second Nature*, no. 3 (2010): 138-163.

Goodfellow, Ian, Jean Pouget-Abadie, Mehdi Mirza, Bing Xu, David Warde-Farley, Sherjil Ozair, Aaron Courville and Yoshua Bengio. "Generative Adversarial Nets". *Advances in Neural Information Processing Systems (NIPS)*, vol. 27 (2014): 2672-80.

Instaloader. "Download Instagram Photos and Metadata". *Instaloader* (n.d.). Accessed: 22 March, 2023. https://instaloader.github.io/

Karras, Tero, Samuli Laine and Timo Aila. "A Style-Based Generator Architecture for Generative Adversarial Networks". *2019 IEEE/CVF Conference on Computer Vision and Pattern Recognition (CVPR). IEE* (2019): 4396-4405. DOI: https://doi.org/10.1109/CVPR.2019.00453

Klingemann, Mario. "Mario Klingemann, Artist Working with Code, AI and Data". *Quasimondo* (2019).

Kurdi, Benedek, Shayn Lozano and Mahzarin R. Banaji. "Introducing the Open Affective Standardized Image Set (OASIS)." *Behavior Research Methods*, vol. 49, no. 2 (2017): 457-470. DOI: https://doi.org/10.3758/s13428-016-0715-3

Kwastek, Katja. *Aesthetics of Interaction in Digital Art*. MIT Press, 2013. DOI: https://doi.org/10.7551/mitpress/9592.001.0001

Lang, Peter J., Margaret M. Bradley and Bruce N. Cuthbert. *International Affective Picture System (IAPS)*. NIMH, Center for the Study of Emotion & Attention, 2005. DOI: https://doi.org/10.1037/t66667-000

Lara-Cabrera, Raúl and David Camacho. "A taxonomy and state of the art revision on affective games." *Future Generation Computer Systems*, vol. 92 (2019, March): 516-525. DOI: https://doi.org/10.1016/j.future.2017.12.056

Machajdik, Jana and Allan Hanbury. "Affective Image Classification Using Features Inspired by Psychology and Art Theory." *Proceedings of the International Conference on Multimedia - MM '10*, (2010): 83-92. New York: ACM Press. DOI: https://doi.org/10.1145/1873951.1873965

Marchewka, Artur, Łukasz Żurawski, Katarzyna Jednoróg and Anna Grabowska. "The Nencki Affective Picture System (NAPS): Introduction to a novel, standardized, wide-range, high-quality, realistic picture database." *Behavior Research Methods*, vol. 46, no. 2 (2014): 596-610. DOI: https://doi.org/10.3758/s13428-013-0379-1

Mikels, Joseph A., Barbara L. Fredrickson, Gregory R. Larkin, Casey M. Lindberg, Sam J. Maglio and Patricia A. Reuter-Lorenz. "Emotional category data on images from the international affective picture system." *Behavior Research Methods*, vol. 37, no. 4 (2005): 626-630. DOI: https://doi.org/10.3758/BF03192732

Mokady, Ron, Amir Hertz and Amit H. Bermano. "ClipCap: CLIP Prefix for Image Captioning". *ArXiv* (2021, November). DOI: https://doi.org/10.48550/arXiv.2111.09734

Park, Minho, Hak Gu Kim and Yong Man Ro. "Photo-Realistic Facial Emotion Synthesis Using Multi-Level Critic Networks with Multi-Level Generative Model". *Lecture Notes in Computer Science (Including Subseries Lecture Notes in Artificial Intelligence and Lecture Notes in Bioinformatics)*, vol. 11296 (2019): 3-15. Cham: Springer. DOI: https://doi.org/10.1007/978-3-030-05716-9_1

Radford, Alec, Luke Metz and Soumith Chintala. "Unsupervised Representation Learning with Deep Convolutional Generative Adversarial Networks". *ArXiv* (2015, November). DOI: https://doi.org/10.48550/arXiv.1511.06434

Ridler, Anna. "Repeating and Mistranslating: The Associations of GANs in an Art Context". *Machine Learning for Creativity and Design, NIPS 2017 Workshop* (2017).

Sharma, Piyush, Nan Ding, Sebastian Goodman and Radu Soricut. "Conceptual Captions: A Cleaned, Hypernymed, Image Alt-Text Dataset For Automatic Image Captioning". *ACL 2018 - 56th Annual Meeting of the Association for Computational Linguistics, Proceedings of the Conference (Long Papers)*, vol. 1 (2018): 2556-2565. DOI: https://doi.org/10.18653/V1/P18-1238

Wang, Junpeng. "Interpreting and Diagnosing Deep Learning Models: A Visual Analytics Approach". The Ohio State University, 2019. http://rave.ohiolink.edu/etdc/view?acc_num=osu1555499299957829

Williams, Duncan, Alexis Kirke, Eduardo R. Miranda, Etienne Roesch, Ian Daly and Slawomir Nasuto. "Investigating affect in algorithmic composition systems". *Psychology of Music*, vol. 43, no. 6 (2015): 831-854. DOI: https://doi.org/10.1177/0305735614543282

Xu, Feiyu, Hans Uszkoreit, Yangzhou Du, Wei Fan, Dongyan Zhao and Jun Zhu. "Explainable AI: A Brief Survey on History, Research Areas, Approaches and Challenges". *Lecture Notes in Computer Science (Including Subseries Lecture Notes in Artificial Intelligence and Lecture Notes in Bioinformatics)*, vol. 11839 (2019): 563-574. Cham: Springer. DOI: https://doi.org/10.1007/978-3-030-32236-6_51

You, Quanzeng, Jiebo Luo, Hailin Jin and Jianchao Yang. "Building a Large Scale Dataset for Image Emotion Recognition: The Fine Print and The Benchmark." *Proceedings of the AAAI Conference on Artificial Intelligence*, vol. 30, no. 1 (2016). DOI: https://doi.org/10.1609/aaai.v30i1.9987

## CV

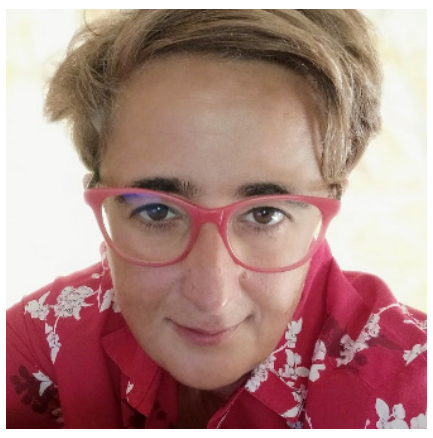

**Amalia Foka**
University of Ioannina
afoka@uoi.gr

A Greek creative Artificial Intelligence researcher and educator who explores the intersection of computer science and art. Her work makes use of different artificial intelligence technologies and social media data to generate and study art. Her work has been presented and published internationally in publications such as the *Leonardo Journal* (MIT Press), WRO Media Art Biennale, ISEA, EVA London and many more. She is currently an Assistant Professor in Computer Science: Multimedia Applications for the Arts at the School of Fine Arts, University of Ioannina, Greece, where she teaches courses on creative coding, creative interactive systems, and generative AI art. She holds a BEng in Computer Systems Engineering (1998) and an MSc in Advanced Control (1999) from the University of Manchester Institute of Science & Technology (UMIST) in the United Kingdom. She also holds a PhD in Robotics (2005) from the Department of Computer Science of the University of Crete.

UOC Universitat Oberta de Catalunya